\documentclass[runningheads]{llncs}
\usepackage[T1]{fontenc}
\usepackage{graphicx,verbatim}
\usepackage{booktabs}
\usepackage{xcolor}
\usepackage{amsmath}
\usepackage{hyperref}
\usepackage[square,numbers,sort&compress,sectionbib]{natbib}
\usepackage{xspace}

\makeatletter
\DeclareRobustCommand\onedot{\futurelet\@let@token\@onedot}
\def\@onedot{\ifx\@let@token.\else.\null\fi\xspace}
\def\eg{\emph{e.g}\onedot}

\begin{document}
\title{A 4D Representation for Training-Free Agentic Reasoning from Monocular Laparoscopic Video}

%

\author{Maximilian Fehrentz*\inst{1,2,4} \and
Nicolas Stellwag*\inst{2} \and
Robert Wiebe\inst{2} \and
Nicole Thorisch\inst{2} \and
Fabian Grob\inst{2} \and
Patrick Remerscheid\inst{2} \and
Ken-Joel Simmoteit\inst{2} \and
Benjamin D. Killeen\inst{1,4} \and
Christian Heiliger\inst{3} \and
Nassir Navab\inst{1,4}}
\authorrunning{M. Fehrentz et al.}
%
\institute{Computer Aided Medical Procedures, TU Munich, Munich, Germany \and
TUM.ai, Munich, Germany \and
Hospital of the LMU Munich, Ludwig-Maximilians-Universität (LMU), Munich, Germany \and
Munich Center for Machine Learning, Munich, Germany}

\maketitle              
\begin{abstract}
Spatiotemporal reasoning is a fundamental capability for artificial intelligence (AI) in soft tissue surgery, paving the way for intelligent assistive systems and autonomous robotics.
While 2D vision-language models show increasing promise at understanding surgical video, the spatial complexity of surgical scenes suggests that reasoning systems may benefit from explicit 4D representations.
Here, we propose a framework for equipping surgical agents with spatiotemporal tools based on an explicit 4D representation, enabling AI systems to ground their natural language reasoning in both time and 3D space.
Leveraging models for point tracking, depth, and segmentation, we develop a coherent 4D model with spatiotemporally consistent tool and tissue semantics.
A Multimodal Large Language Model (MLLM) then acts as an agent on tools derived from the $explicit$ 4D representation (e.g., trajectories) without any fine-tuning.
We evaluate our method on a new dataset of 134 clinically relevant questions and find that the combination of a general purpose reasoning backbone and our 4D representation significantly improves spatiotemporal understanding and allows for 4D grounding.
We demonstrate that spatiotemporal intelligence can be ``assembled'' from 2D MLLMs and 3D computer vision models without additional training. Code, data, and examples are available at \href{https://tum-ai.github.io/surg4d/}{https://tum-ai.github.io/surg4d/}.


\end{abstract}

\section{Introduction}


Tool use is rapidly becoming a cornerstone for artificial intelligence (AI) agents, enabling them to form explicit interfaces with the world and solve complex tasks in a training-free manner~\cite{schick2023toolformer}. In robot-assisted surgery, systems that can understand and reason about the procedure have the potential to unlock a new level of computer assistance~\cite{fehrentz2025bridgesplat}, skill assessment~\cite{bitner2024kinematic}, and autonomy~\cite{kim2025srt}, thereby improving the quality and accessibility of care.
However, the exact tools and representations required to achieve this remain unclear. Here, we take inspiration from the surgical cognitive process, which relies heavily on 3D spatial and temporal understanding to identify and expose anatomical dissection planes~\cite{connor2014using}, and develop 4D spatiotemporal tools to augment AI agents. Despite the importance of these operative principles in current practice, 4D spatiotemporal tools and representations remain largely unexplored in surgical AI.

\begin{figure}[t]
    \centering
    \includegraphics[width=1.0\linewidth]{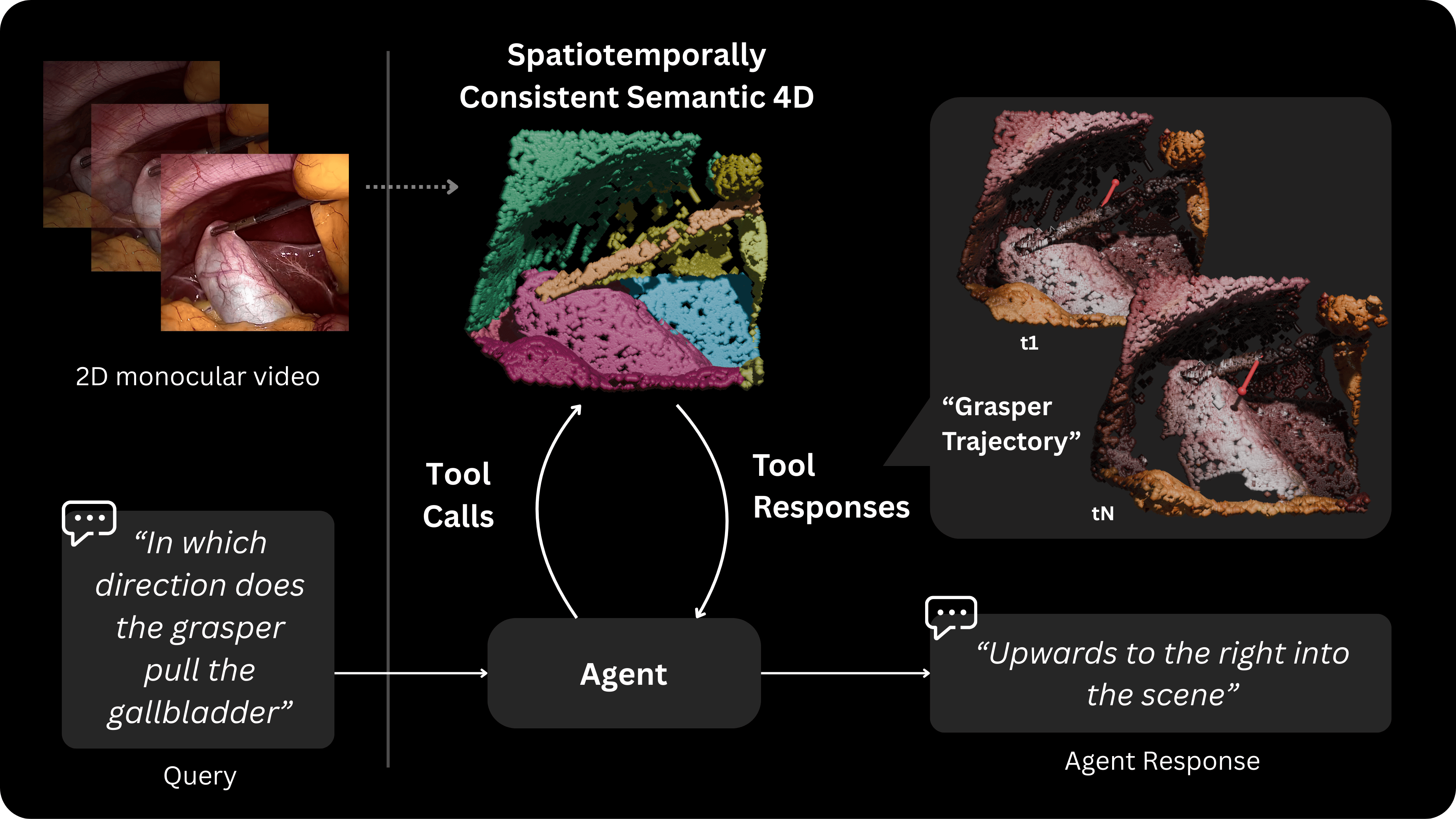}
    \caption{Given a monocular 2D video sequence, our method creates a tracked semantic 4D representation and provides spatiotemporal information as tools to a 2D MLLM Agent. Upon receiving a query, the agent iteratively requests and interprets spatiotemporal information from our 4D representation to inform its response.}
    \label{fig:method}
\end{figure}

Until now, surgical AI has made significant progress by reasoning in 2D image space. Spatiotemporal representations, such as surgical scene graphs~\cite{shin2025towards,murali2023encoding} or frame-level action triplets~\cite{nwoye2022rendezvous}, have been used to model interactions between anatomical structures and surgical tools.
Vision-language models, on the other hand, ingest tokens extracted from 2D image patches, enabling downstream reasoning tasks like 2D visual question answering (VQA) with a transformer model~\cite{ssq-vqa,surgical-vqa}.
Multimodal Large Language Models (MLLMs) go one step further in grounding reasoning based on 2D bounding boxes~\cite{wang2025endochat,wang2024surgical,zeng2025surgvlm} or with coarse language descriptions~\cite{chen2025surgllm}. Recent efforts have also trained MLLMs from scratch across surgical disciplines~\cite{yang2026large}, but remain constrained to action and triplet detection.
Beyond 2D, SurgTPGS~\cite{huang2025surgtpgs} introduces a language-aligned 3D representation by embedding language features into 3D Gaussians. While this enables language-based queries on 3D space, significant hurdles remain for agentic reasoning on that representation due to the contrastive training and lack of consistent, temporal information.

Reasoning on 4D representations in surgery has remained elusive for several reasons, including the fact that 4D reconstruction on 2D monocular laparoscopic video suffers from adverse visual features, \eg, specular highlights, and little camera motion.
Recently, though, computer vision foundation models have crossed a critical threshold. Depth Anything 3~\cite{da3} in combination with Cotracker 3~\cite{cotracker3} supports 4D reconstruction, camera parameter estimation, and tracking that is sufficiently reliable even in the surgical setting. Simultaneously, agentic tool calling has opened a door for providing \textit{training-free} spatial interfaces to reasoning models~\cite{schick2023toolformer}.
With these capabilities, we introduce a 4D representation with spatially and temporally consistent semantics that supports agentic reasoning with tools.
In evaluating this approach, we are motivated by the lack of surgical AI benchmarks specifically targeting tissue handling, that is, the identification of 3D grasping points and manipulation in specific directions.
Thus, with Qwen3-VL as the agent ~\cite{bai2025qwen3}, we evaluate our framework on the novel task of 4D grounding, using a subset of the Cholec-80 dataset~\cite{cholec80} annotated with 134 clinically relevant queries about when and where interactions occur in 3D space, and in what direction tissue is being manipulated.


\section{Method}

\subsection{Semantic Tracked 4D from Monocular Video}
Given a monocular video from laparoscopic surgery, we estimate depth and camera parameters using Depth Anything 3 (DA3) \cite{da3} and extract semantic segmentation masks with SASVi \cite{sivakumar2025sasvi}. To perform 4D reasoning, we must identify and track instances in 4D. However, generally, depth models create unrelated 3D point clouds across timesteps, and segmentation models do not provide instance re-identification over time. Therefore, we also obtain 2D tracking from Cotracker 3 \cite{cotracker3}. By lifting the tracked points using depth and camera parameters from DA3, we obtain a tracked but sparse 4D \textit{control} point cloud $\mathcal{P}_\text{control}$. Since \textit{dense} point clouds are desirable for higher resolution, pixels are associated with their K-nearest control points and are assigned 4D positions by interpolation, yielding $\mathcal{P}_\text{dense}$. To enrich the resulting tracked 4D point cloud with semantics to obtain $\mathcal{P}_\text{semantic}$, we first convert the frame-wise semantic segmentations provided by SASVi to instance segmentations, using connected components.
Instance labels from the set of initial frames are lifted to 3D and tracked
to a reference timestep $t_\text{ref}$. At $t_\text{ref}$, we consider all pairs of instances that stem from different initial frames and have the same semantic class.
Each considered pair should be merged if its 3D containment score is above a threshold.
A union-find operation yields the final set of merged, time-persistent 3D instances $\mathcal{P}_\text{semantic}$.
Importantly, the proposed merging approach allows for new instances to emerge or existing ones to disappear, like surgical tools entering or leaving the scene. We propose several intelligent mechanisms to enhance the quality of the representation.

\noindent\textbf{Control Point Assignment in 3D}: When naively associating pixels with 2D Cotracker points, the lack of depth can cause pixels to get assigned to control points that track both background and foreground objects. Since 4D positions are then computed by interpolating the assinged control points' positions, those points end up floating between foreground and background. We circumvent this by unprojecting pixels first, and only then assigning them to their nearest control points in 3D.

\noindent\textbf{Background/Foreground Jump Filtering}: The method is initially susceptible to 2D tracking errors. Small tracking inaccuracies can lead to a pixel moving between background tissue and surgical tool in the foreground. When naively lifting the 2D tracking control points, they jump back and forth across large depth deltas. Therefore, we remove points that undergo large, abrupt changes along the z-axis.

\noindent\textbf{Depth Maintenance}: If tracking is accurate, a point located on background anatomy should remain there during tool occlusion. However, when naively lifting points to 3D for that frame, that occluded point would receive the depth of the occluding object and jump into the foreground. Since Cotracker 3 also provides visibility per point, we ``forward fill'' depth values. If a tracked point becomes invisible (occluded or out of view), we do not overwrite its depth with new observations. Instead, we maintain its depth from the last time it was observed. Therefore, occluded and out-of-view points will remain at their last observed position. We consider this behavior advantageous, as we can re-identify them on recurrence and use them for context, while knowing they are not to be trusted as an exact representation.

\noindent\textbf{Forward/Backward Tracking}: Cotracker is usually initialized from a single frame and keeps track of those initial points. While it appears sensible to initialize several tracking views, and it is generally needed for longer sequences, the focus of this work is on short interactions. Therefore, it is advantageous to maintain a single, clean tracking instance initialized from the middle frame of a clip. This has the highest likelihood of containing the most relevant points, and tracks both backward and forward in time.

\noindent\textbf{Depth Gradient Filtering}: We observe that DA3 produces oversmooth edges along objects, leading to tool boundaries ``fading out'' rather than constituting a sharp edge. We therefore introduce depth gradient filtering into the proposed method to obtain cleaner object boundaries.

\subsection{Agentic Reasoning \& Spatiotemporal Tools}

We package our 4D representation as an agentic toolkit that can be used with any
MLLM backbone. When prompted, the resulting agent receives a set of nodes with semantic classes and coarse geometric information such as means.
Based on this, it can start exploring the 4D scene with spatiotemporal tools
that compute minimum distances, 3D overlap scores, 3D overlap positions, and
relative motion between semantic instances over time, as well as 4D trajectories
per instance. Since video and 4D geometry carry complementary information, we
additionally propose an agent with frame fetching ability that can request video
frames at individual timesteps using another tool. This can be useful, for example, to disambiguate whether an L-hook is just poking tissue or coagulating.


\section{Experiments \& Results}


\subsubsection{Data \& Implementation Details}
A board-certified surgeon annotated 134 clinically relevant queries across 25 short clips of 3-4 seconds from Cholec80~\cite{cholec80} that show tool-tissue interaction. The clips have a framerate of 25 fps and are annotated at timesteps of frame stride 4. The queries are split into three categories: 46 spatial queries, 47 temporal queries, and 41 directional queries. Importantly, \textbf{each query} requires spatial or spatiotemporal reasoning, as they all concern spatial interactions. Spatial queries ground in 3D for a single timestep, temporal queries ground in time, and directional queries determine a dominant, discrete 3D direction for an interaction over time. Across all tasks, we compare our agent against the same 2D MLLM with image and video input as a baseline. 2D segmentations are provided to the baseline with a variation of Set-of-Mark prompting~\cite{som}, where we also provide the semantic class. All experiments run on a single NVIDIA H200. We use the DA3NESTED-GIANT-LARGE-1.1 version of the DA3 model because it estimates metric depth. When using Qwen3-VL in its 8B configuration, the method can run on a single GPU with 48 GB VRAM. Our MLLM backbone is Qwen3-VL-32B-Thinking-FP8~\cite{bai2025qwen3} served by vLLM~\cite{vllm} with default sampling configuration.

\subsubsection{Spatial}
\begin{figure}[t]
    \centering
    \includegraphics[width=1.0\linewidth]{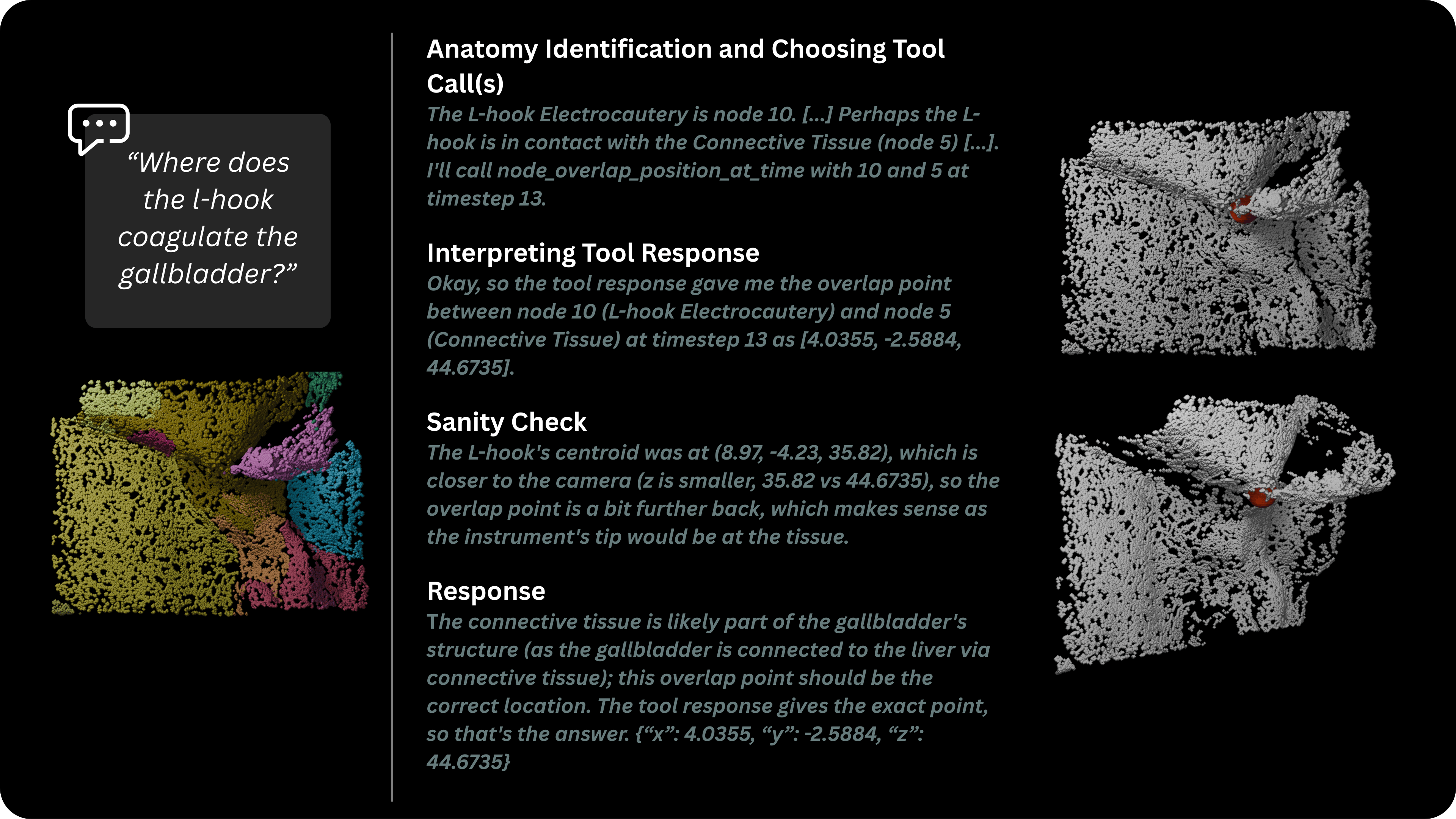}
    \caption{Example result and partial reasoning trace on a spatial query. Despite not having access to a specific gallbladder semantic instance, the agent understands that the connective tissue class is the desired connection between gallbladder and liver and localizes the interaction correctly.}
    \label{fig:spatial}
\end{figure}

\begin{table}[t]
    \centering
    \caption{Spatial, temporal point-in-time, temporal interval, and directional grounding.}
    \label{tab:main_table}
    \newcommand{\mstd}[2]{$#1_{\scriptsize\pm\mathit{#2}}$}
    \renewcommand{\arraystretch}{1.2}
    \begin{tabular*}{\textwidth}{@{\extracolsep{\fill}}llrrrr@{}}
        \toprule
        && \multicolumn{1}{c}{Spatial} & \multicolumn{1}{c}{Temp. PIT} & \multicolumn{1}{c}{Temp. Interval} & \multicolumn{1}{c}{Directional} \\
        \cmidrule(l){3-3} \cmidrule(l){4-4} \cmidrule(l){5-5} \cmidrule(l){6-6}
        ~ & Model & \multicolumn{1}{r}{L2 (px) $\downarrow$} & (timesteps) $\downarrow$ & IoU $\uparrow$ & \multicolumn{1}{r}{L1 $\downarrow$ } \\
        \midrule
        \multicolumn{2}{l}{\textit{Baselines}}\\
        & MLLM & \mstd{90.76}{65.08} & \mstd{6.56}{5.05} & \mstd{\textbf{0.51}}{0.34} & \mstd{1.15}{0.88} \\
        & MLLM w/ SoM & \mstd{142.77}{73.62} & \mstd{5.12}{3.39} & \mstd{\textbf{0.51}}{0.31} & \mstd{0.83}{0.85} \\
        \multicolumn{2}{l}{\textit{Ours}}\\
        & Agent & \mstd{\textbf{46.54}}{53.40} & \mstd{\textbf{3.62}}{3.79} & \mstd{0.39}{0.31} & \mstd{0.18}{0.54} \\
        & Agent w/ Frame Fetching & \mstd{50.87}{61.38} & \mstd{4.12}{4.40} & \mstd{0.39}{0.30} & \mstd{\textbf{0.17}}{0.53} \\
        \multicolumn{2}{l}{\textit{Ablations}}\\
        & No Jump Filter & \mstd{63.37}{74.65} & \mstd{5.19}{4.75} & \mstd{0.36}{0.34} & \mstd{0.22}{0.61} \\
        & No Depth Maintenance & \mstd{48.34}{50.02} & \mstd{8.19}{5.95} & \mstd{0.32}{0.28} & \mstd{0.25}{0.62} \\
        & Multi-frame Tracking & \mstd{49.35}{51.91} & \mstd{7.50}{5.32} & \mstd{0.34}{0.29} & \mstd{0.28}{0.65} \\
        & Qwen3-VL-8B-Thinking & \mstd{61.89}{70.33} & \mstd{6.25}{5.24} & \mstd{0.28}{0.33} & \mstd{0.85}{0.98} \\
        \bottomrule
    \end{tabular*}
    \renewcommand{\arraystretch}{1.0}
\end{table}

To evaluate spatial queries, the suggested method returns 3D points. The 2D baseline, however, can only return a pixel prediction in image space. Therefore, ground truth is provided as 2D pixel annotations. Errors are computed as L2 distance in screen space. For 3D predictions, the error is computed after projection to the image plane. Note that although both approaches yield the same error metric, \textbf{the spatiotemporal model is always solving the more difficult task} by grounding in (estimated) metric 3D.

Results are reported in Tab.~\ref{tab:main_table}. The proposed method significantly outperforms the best 2D baseline, improving by almost 50\% while locating the query in 3D rather than 2D.
Surprisingly, providing the semantics to the baseline inhibits its grounding capabilities.
The proposed method achieves the lowest mean error by reasoning \textit{exclusively} on the spatiotemporal tools. However, standard deviations are high across all approaches, indicating that the methods either locate the queries very well or misinterpret significantly. The difference in mean between tools with and without vision encoder is small relative to the observed variability. However, given the baseline's poor performance, the vision encoder's input during the reasoning process may be detrimental and conflicting with tool responses. 
A 3D example with its partial reasoning trace is shown in Fig.~\ref{fig:spatial}.

\subsubsection{Temporal}
\begin{figure}[t]
    \centering
    \includegraphics[width=\linewidth]{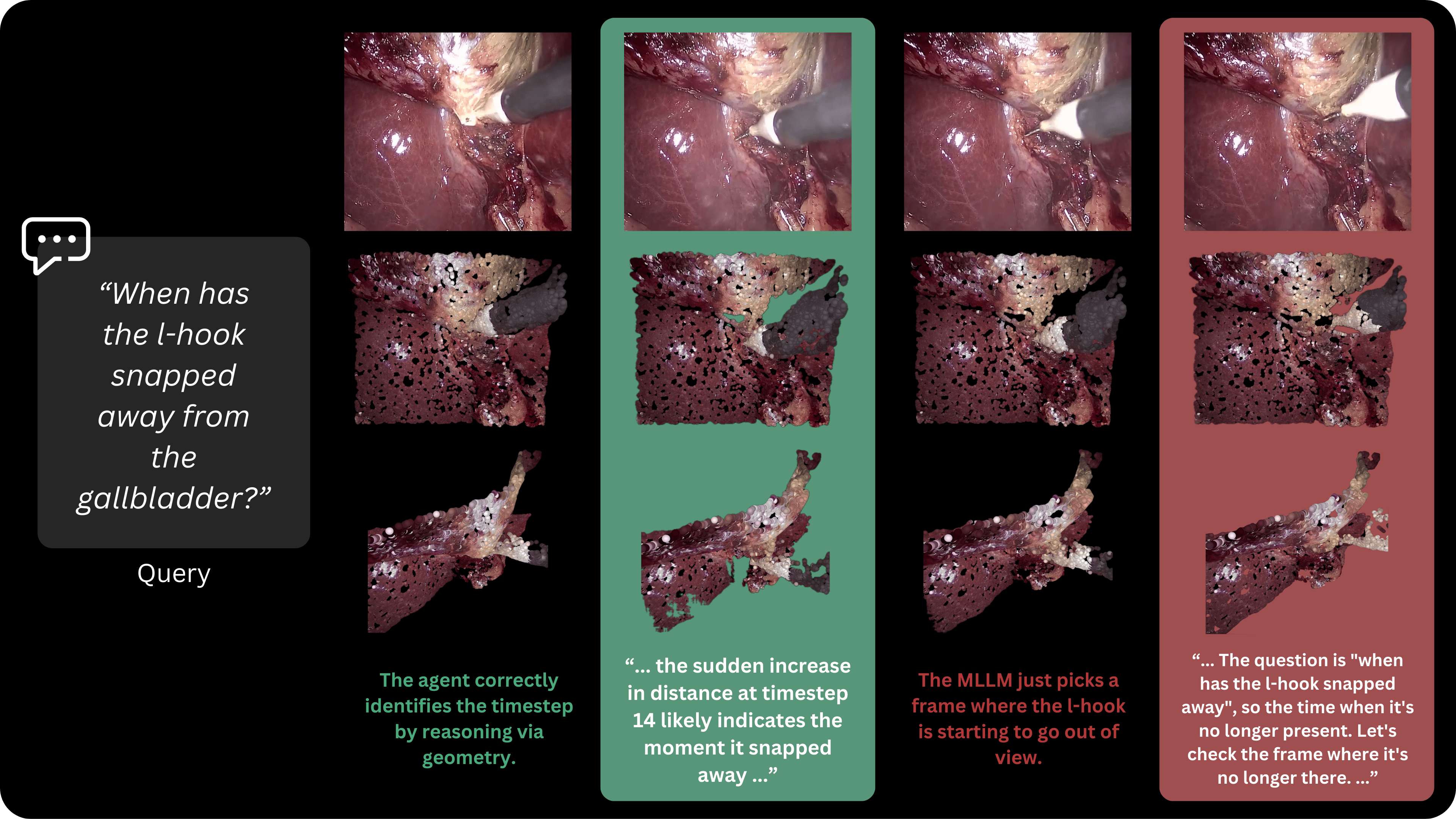}
    \caption{Temporal example showing 2D video, 4D from two perspectives, and the derived model decisions}
    \label{fig:temporal}
\end{figure}
Temporal queries are divided into two types: \emph{Point-in-time} (PIT) queries
ask to localize an event happening at a single timestep. \emph{Interval} queries
require predicting one or multiple ranges of frames. We compute the PIT error as
the absolute difference between the predicted and ground truth timestep. For
interval queries, we compute the Intersection over Union (IoU) between predicted temporal
ranges and ground truth.

An example of PIT query with responses from both baseline and proposed method is
shown in Fig.~\ref{fig:temporal}. Quantitative results are reported in
Tab.~\ref{tab:main_table}. For PIT queries, unlike spatial queries, providing
semantics via Set-of-Mark (SoM) prompting helps the 2D baseline. It has no
effect on interval queries. Compared to the baselines, our agent is worse at
predicting intervals but better at localizing points in time, independent of
frame fetching ability. We hypothesize that the approaches have different
strengths and weaknesses. For 2D baselines, accurately determining 3D
interaction is highly ambiguous, but visual cues from deforming tissue help.
Spatiotemporal tools simplify this task, given the underlying point cloud is
accurate enough. Since reconstructions are based on monocular video without
camera parameters, they are still noisy, as shown in Fig.~\ref{fig:temporal},
which can mislead. Models occasionally fail to give a valid, parseable response
to temporal queries. In this case, we set the PIT error to the scene's number of
timesteps or assign an IoU of 0. Overall, all evaluated methods struggle with
temporal queries.


\subsubsection{Directional}
Directional queries seek to determine the dominant direction of an interaction over time.
Similar to spatial tasks, we pick the prediction format such that comparisons against the 2D baselines are fair.
Since Qwen3-VL was trained on predicting 3D bounding boxes, it is a fair assumption that the model has an internal, implicit notion of depth.
However, contrary to the spatiotemporal tools, it does not maintain an explicit 3D coordinate system.
Therefore, we ask the models to respond with a discrete 3D unit direction (-1, 0, or 1) along all three axes of the world coordinate system.
Errors are computed as the mean absolute error between predicted and ground truth direction.
As annotating directions in 2D is difficult even for expert surgeons, axes considered neutral or ambiguous are annotated as 0 and not considered for the error computation.
When models fail to give a parsable response, we assign the maximal error of 2.

Results are shown in Tab.~\ref{tab:main_table}.
For the 2D baseline, similar to temporal queries, Set-of-Mark prompting helps 2D models to determine direction as it may enhance the model's ability to implicitly reidentify and track objects.
The agent improves over the best baseline by 79\% and significantly benefits from its ability to obtain 4D trajectories. Access to the frame fetching tool does not lead to significant differences.

\subsubsection{Ablations}
We ablate the key components of the proposed 4D representation by removing jump filtering (mitigating 2D tracking errors), depth maintenance (forwarding depth for occluded or out-of-view points), and replacing single-view tracking with multi-view initialization. In the latter, tracking is initialized from the first (forward), middle (forward and backward), and last (backward) frames. Since tool-calling and reasoning typically scale with model size, we additionally compare Qwen3-VL Thinking with 8B and 32B parameters. Results are reported in Tab.~\ref{tab:main_table}. Contrary to intuition, multi-view initialization does not improve performance. We hypothesize that independently tracked point clouds introduce noisy control points, whose errors accumulate during interpolation. Removing jump filtering degrades performance, particularly on temporal tasks, as foreground-background motion amplifies tracking noise. Omitting depth maintenance harms all tasks since occluded background points are projected into the foreground, preserving geometric continuity at single timesteps but introducing semantic inconsistencies over time. As expected, Qwen3-VL with 8B parameters is significantly worse, mostly due to its difficulty parsing tool answers correctly and adhering to the desired response formatting.

\section{Discussion \& Conclusion}
Providing tools as an interface raises the general question of how to \textit{represent} spatiotemporal information.
Currently, there is no consensus on whether \textit{explicit} 3D / 4D representations are necessary.
There is no fundamental reason why a ``2D model'' receiving depth and tracking cannot build such a representation internally.
However, in this work, we adopt an \textit{explicit} spatiotemporal representation for its interpretability and downstream integration advantages with registered preoperative imaging.
Although a 2D-compatible benchmark on 2D-only baselines allows us to demonstrate the concept, more fine-grained tasks will necessitate 4D annotations on metric ground truth in the future. In a similar vein, while Qwen3-VL serves as an effective backbone to show feasibility and effectiveness, other surgical MLLMs may interact with spatiotemporal representations in comparatively beneficial ways.
Lastly, the suggested method leverages DA3, Cotracker 3, and SASVi as offline methods, which benefit considerably from the full temporal context. This is currently needed, considering the difficulty of obtaining tracked semantic 4D from monocular video without camera parameters, but all methods can potentially run online at frame rates that might enable online intraoperative use as well. As a first step, we have introduced a tracked semantic 4D representation from monocular laparoscopic video without camera parameters that agents can interact with via spatiotemporal tools to tackle the clinically relevant task of understanding 4D interactions in the abdomen.

\bibliographystyle{splncs04}
\bibliography{mybibliography}

\end{document}